\pdfoutput=1

\documentclass[11pt]{article}

\usepackage[final]{acl}

\usepackage{times}
\usepackage{latexsym}

\usepackage[T1]{fontenc}

\usepackage[utf8]{inputenc}

\usepackage{microtype}

\usepackage{inconsolata}

\usepackage{graphicx}

\usepackage{multirow}
\usepackage{multicol}
\usepackage{booktabs}
\usepackage{subcaption}

\usepackage{graphicx}
\usepackage{colortbl}
\usepackage[most]{tcolorbox}

\tcbset{
  parskip/.style={
    before={\par\pagebreak[0]\parindent=0pt},
    after={\parfillskip=0pt\par},
  },
}
\newtcolorbox{resultbox}[1][]{%
    colback=black!3,
    colframe=black!3,
    notitle,
    sharp corners,
    borderline west={2pt}{0pt}{gray!80!black},
    enhanced,
    breakable,
    boxsep=0pt,
    left=4pt,right=2pt,top=2pt,bottom=2pt,
    }

\newcommand\blfootnote[1]{%
  \begingroup
  \renewcommand\thefootnote{}\footnote{#1}%
  \addtocounter{footnote}{-1}%
  \endgroup
}

\title{Exploring the Performance of Large Language Models on \\ Subjective Span Identification Tasks}

\author{Alphaeus Dmonte\textsuperscript{1}, Roland Oruche\textsuperscript{2}, Tharindu Ranasinghe\textsuperscript{3} \\ \vspace{2mm} {\bf Marcos Zampieri\textsuperscript{1}, Prasad Calyam\textsuperscript{2}} \\ 
  \textsuperscript{1}George Mason University, VA, USA 
  \textsuperscript{2}University of Missouri, MO, USA \\
 \textsuperscript{3}Lancaster University, UK \\
  \texttt{admonte@gmu.edu} \\
  }

\begin{document}
\maketitle
\begin{abstract}
     Identifying relevant text spans is important for several downstream tasks in NLP, as it contributes to model explainability. While most span identification approaches rely on relatively smaller pre-trained language models like BERT, a few recent approaches have leveraged the latest generation of Large Language Models (LLMs) for the task. Current work has focused on explicit span identification like Named Entity Recognition (NER), while more subjective span identification with LLMs in tasks like Aspect-based Sentiment Analysis (ABSA) has been underexplored. In this paper, we fill this important gap by presenting an evaluation of the performance of various LLMs on text span identification in three popular tasks, namely sentiment analysis, offensive language identification, and claim verification. We explore several LLM strategies like instruction tuning, in-context learning, and chain of thought. Our results indicate underlying relationships within text aid LLMs in identifying precise text spans.
     \blfootnote{WARNING: This paper contains examples that are offensive in nature.}
\end{abstract}






\section{Introduction}\label{sec:introduction}

Offensive language identification, sentiment analysis, and claim verification are some of the most widely studied tasks at the intersection of social media analysis and NLP \cite{sandu2024bibliometric}. Most of the research on these tasks focuses on predicting post-level categorical labels. In the case of sentiment analysis, for example, these are often expressed in terms of {\em positive}, {\em neutral}, and {\em negative} labels or a Likert-scale representing the positive to negative continuum \cite{birjali2021comprehensive}. 

Various studies have addressed model explainability by developing frameworks, datasets, and models to identify attributes in texts through token span prediction. For example, in the toxic spans detection task, models predict the spans of toxic posts that are indicative of toxic label prediction \cite{pavlopoulos2021semeval, mathew2021hatexplain}. Going beyond independent token spans, researchers have also proposed more structured formulations to capture relationship between textual elements. One of the most well-established of these formulations is Aspect-Based Sentiment Analysis (ABSA) \cite{pontiki-etal-2014-semeval, pontiki-etal-2015-semeval}, which aims to detect aspects and their associated sentiments within a text. This approach is particularly effective for cases with mixed sentiments, such as \emph{“The food was delicious, but the service was extremely slow”} in a restaurant review. In this example different parts of the text express opposing opinions. In the same vein, in this paper we consider both complex and simple texts and define them as follows: 

\begin{resultbox}
\textbf{Complex Text} - A text containing more than one type of interrelated spans, and these related spans belong to different categories, such as \textsc{Target} and \textsc{Aspect} in ABSA.\\

\textbf{Simple Text} - A text with only one span category such as a {\em toxic span} or {\em claim span} containing a toxic expression and a claim respectively.
   
\end{resultbox}

\noindent LLMs have achieved state-of-the-art performance across various NLP tasks, including generation and prediction \cite{minaee2024large}. Recent studies on evaluating LLMs for sequence labeling tasks such as Named Entity Recognition (NER) \cite{wang2023gpt, pang2023guideline} suggest that BERT models still outperform LLMs in the in-context learning setting. \citet{li2023label} and \citet{dukic2024looking} proposed approaches that transform the objective of LLMs to improve their performance on classification tasks. While LLMs have been explored for NER and sentiment analysis tasks, they have been unexplored for other token classification tasks like offensive spans and claim spans identification, and our work aims to address this gap.

This paper addresses the following \textbf{research questions}:

\begin{itemize}
    \item {\bf RQ1:} How does the complexity of the text affect the LLM's ability to identify the different types of spans? Do the models identify specific span types more efficiently than others?
    \item {\bf RQ2:} How do the model size and modeling strategies influence the span identification capabilities of LLMs?
    \item {\bf RQ3:} Are LLMs efficient in a low-resource setting?
\end{itemize}






\section{Related Work}

Offensive language identification, sentiment analysis, and claim verification are some of the widely studied text classification tasks. Several datasets with post-level annotations have been released for offensive language~\cite{davidson2017automated, zampieri-etal-2019-predicting, ranasinghe2021mudes, mathew2021hatexplain}, sentiment analysis~\cite{tan2023survey}, as well as claim verification~\cite{wang2017liar, thorne2018fever, schlichtkrull2024averitec}. Most of the approaches to these tasks rely on pre-trained transformer-based language models like BERT~\cite{caselli2020hatebert, sarkar2021fbert, tan2023survey,zhang2023sentiment,dmonte2024evaluation} while, more recently, LLMs have also been explored \cite{pan2023fact,zampieri2023offenseval,dmonte2024claim}.


While most of the work on the aforementioned three tasks addresses post-level analysis, several datasets and approaches for token-level analysis have also been proposed. For offensive language, the TSD \cite{pavlopoulos2021semeval} and HateXplain \cite{mathew2021hatexplain} datasets were introduced to identify the token spans containing offensive or toxic content, or specific rationales contributing to the predicted labels while TBO \cite{zampieri2023target} was created, to identify the offensive spans and their associated targets. 
Similarly, ABSA \cite{pontiki-etal-2014-semeval} aims to identify the aspects, which are token spans from the text describing the targets or entities, as well as the sentiment labels associated with these aspects. \citet{wang2016recursive, wang2017coupled} further annotated this dataset to identify the opinion terms. Approaches like Relation-aware Collaborative Learning (RACL)~\cite{chen2020relation}, which considers the relationship between the different types of spans, have showed promising results. 

While LLMs have not been used extensively for token span identification tasks, there are some works that have leveraged these models on a few similar tasks. \citet{han2023information} leveraged GPT for four tasks, namely NER, Relation Extraction, Entity Extraction, and ABSA. The authors observe that LLMs achieve lower performance compared to smaller BERT-based models. To improve the LLM performance on token spans tasks like NER, ABSA, etc, approaches that remove the causal mask from the LLM layers have been proposed~\cite{li2023label, dukic2024looking}. While these approaches improve the performance on the token spans identification tasks, removing the causal mask changes the training objective of the models, essentially transforming the model from autoregressive to a masked language model. In this work, we leverage the autoregressive capabilities of the LLMs to evaluate their performance using different approaches.

\section{Datasets}
\label{sec:datasets}




We acquire four English datasets for our experiments, two with complex text spans and two with simple text spans. The example instances from each dataset are presented in Table~\ref{tab:dataset-examples}, while the data statistics are presented in Table \ref{tab:data-stats}.

\begin{table}[!ht]
    \centering
    \resizebox{0.45\textwidth}{!}{
    \begin{tabular}{ccccc}
    \toprule
         {\bf Span Type} & {\bf Dataset} & {\bf Train} & {\bf Test} & {\bf Total}  \\
         \midrule
         \multirow{2}{*}{Complex Spans} & TBO & 4,000 & 673 & 4,673 \\
         & ABSA & 3,041 & 800 & 3,841 \\
         \midrule
         \multirow{2}{*}{Simple Spans} & CSI & 3,953 & 362 & 4,315 \\
         & TSD & 8,629 & 2,000 & 10,629 \\
         \bottomrule
    \end{tabular}}
    \caption{Number of Train and Test instances in the datasets used for the experiments.}
    \label{tab:data-stats}
\end{table}


\paragraph{Complex Text Datasets} We acquire {\bf Target Based Offensive Language (TBO)} \cite{zampieri2023target} and the {\bf Aspect Based Sentiment Analysis (ABSA)} dataset by \newcite{pontiki-etal-2014-semeval}. The instances in the TBO dataset were annotated with \emph{Arguments}, which are offensive phrases in the text, and \emph{Target} representing the subject of the arguments. The ABSA dataset is annotated with \emph{Aspects} and their corresponding \emph{Opinion} spans.
 


\paragraph{Simple Text Datasets} We use the {\bf Claim Spans Identification (CSI)}~\cite{mittal2023lost} and {\bf Toxic Span Detection (TSD)}~\cite{pavlopoulos2021semeval} datasets. CSI is annotated with claim spans from social media posts, while the TSD dataset is annotated for toxic and harmful spans.



\begin{table*}[!ht]
    \centering
    \resizebox{0.98\textwidth}{!}{
    \begin{tabular}{ccp{10cm}p{7cm}}
    \toprule
         {\bf Dataset} & {\bf Type} & {\bf Example Instance} & {\bf Annotation}  \\
         \midrule
         \multirow{4}{*}{TBO} & Complex & @USER Time to stop the voter fraud. These people are evil. & Target 1: None, Argument 1: voter fraud, Target 2: these people, Argument 2: are evil \\
         \cmidrule{2-4}
         & Complex & @USER Imma wear my uggs until they turn inside out the hell! & Target 1: @USER, Argument 1: hell \\
         \midrule
         \multirow{4}{*}{ABSA} & Complex & not only was the food outstanding, but the little 'perks' were great & Aspect 1: food, Opinion 1: outstanding, Aspect 2: perks \\
         \cmidrule{2-4}
         & Complex & raga's is a romantic, cozy restaurant & Opinion 1: romantic, Opinion 2: cozy \\
         \midrule
         \multirow{3}{*}{CSI} & Simple & It’s not Rahul Khan, it & Span: It’s not Rahul Khan, it \\
         \cmidrule{2-4}
         & Simple & They will try everything to steal it. We will not let them! & Span: They will try everything to steal it. \\
         \midrule
         \multirow{4}{*}{TSD} & Simple & 'Another violent and aggressive immigrant killing a innocent and intelligent US Citizen.... Sarcasm' & Argument 1: violent and aggressive immigrant \\
         \cmidrule{2-4}
         & Simple & What a knucklehead. How can anyone not know this would be offensive?? & Argument 1: knucklehead \\

         \bottomrule
    \end{tabular}}
    \caption{Example instances from each dataset. The instances, along with their respective text span annotations, are shown. Complex text (TBO and ABSA) have two types of spans, while simple text (CSI and TSD) have only one span type.}
    \label{tab:dataset-examples}
\end{table*}

\section{Experiments}

We describe the models used in our experiments. BERT models are fine-tuned with task-specific datasets, while instruction-tuning, in-context learning, and chain-of-thought are used for LLMs.

\paragraph{Baselines} We use the BERT-large~\cite{devlin-etal-2019-bert} model as a baseline for our experiments. We fine-tune the model with the task-specific training datasets. For the progress test, the models are fine-tuned with a randomly sampled subset of the training set.

\paragraph{LLMs} We utilize the Qwen2.5~\cite{yang2024qwen2} and Llama-3.1~\cite{llama3modelcard} model families due to the availability of multiple model sizes, enabling evaluation across different model scales. Specifically, we employ the 7B, 14B, 32B, and 72B parameter variants for Qwen, and 8B and 70B for Llama.

\paragraph{Approaches} We utilize three LLM approaches in our experiments. All LLMs are instruction-tuned (IT) on all the tasks. The task-specific example prompts are shown in Appendix~\ref{sec:inst-prompts}. In-context Learning (ICL) is used to evaluate off-the-shelf models. More specifically, 0-, 3-, and 5-shot approaches are used. For few-shot learning, an embedding for each test instance is generated using a sentence-transformer model, and top-{\em k} similar instances from the training set are used as few-shot exemplars. Finally, we employ the zero-shot chain-of-thought (CoT)~\cite{kojima2022large} prompting strategy for token spans identification.




\begin{table*}[!ht]
    \centering
    \small
    \resizebox{\textwidth}{!}{
    \begin{tabular}{lcccccccc|cccc}
    \toprule
        \multirow{3}{*}{\bf Model} 
        & \multicolumn{4}{c}{\bf TBO} 
        & \multicolumn{4}{c|}{\bf ABSA}  
        & \multicolumn{2}{c}{\multirow{2}{*}{\bf CSI}} 
        & \multicolumn{2}{c}{\multirow{2}{*}{\bf TSD}} \\
        \cmidrule{2-9}
        & \multicolumn{2}{c}{\bf Target} & \multicolumn{2}{c}{\bf Argument} 
        & \multicolumn{2}{c}{\bf Aspect} & \multicolumn{2}{c|}{\bf Opinion}  
        &  &  &  & \\
        \cmidrule{2-13}
        & TF1 & SF1 & TF1 & SF1 & TF1 & SF1 & TF1 & SF1 
        & TF1 & SF1 & TF1 & SF1 \\
        \midrule
BERT & \cellcolor{teal!76.6}0.766 & \cellcolor{teal!61.1}0.611 & \cellcolor{teal!77.9}0.779 & \cellcolor{teal!60.3}0.603 & \cellcolor{teal!92.4}0.924 & \cellcolor{teal!84.3}0.843 & \cellcolor{teal!90.7}0.907 & \cellcolor{teal!84.0}0.840 & \cellcolor{teal!57.3}0.573 & \cellcolor{teal!12.6}0.126 & \cellcolor{teal!79.4}0.794 & \cellcolor{teal!65.2}0.652 \\
\midrule
Llama-70B-CoT & \cellcolor{teal!50.4}0.504 & \cellcolor{teal!24.4}0.244 & \cellcolor{teal!35.0}0.350 & \cellcolor{teal!9.3}0.093 & \cellcolor{teal!67.0}0.670 & \cellcolor{teal!53.1}0.531 & \cellcolor{teal!54.8}0.548 & \cellcolor{teal!37.6}0.376 & \cellcolor{teal!28.7}0.287 & \cellcolor{teal!7.8}0.078 & \cellcolor{teal!41.8}0.418 & \cellcolor{teal!3.2}0.032 \\
Qwen-72B-CoT & \cellcolor{teal!54.5}0.545 & \cellcolor{teal!31.6}0.316 & \cellcolor{teal!40.4}0.404 & \cellcolor{teal!12.2}0.122 & \cellcolor{teal!76.0}0.760 & \cellcolor{teal!68.2}0.682 & \cellcolor{teal!54.3}0.543 & \cellcolor{teal!43.2}0.432 & \cellcolor{teal!44.4}0.444 & \cellcolor{teal!18.3}0.183 & \cellcolor{teal!31.0}0.310 & \cellcolor{teal!2.8}0.028 \\
\midrule
Llama-70B-0 & \cellcolor{teal!52.3}0.523 & \cellcolor{teal!32.6}0.326 & \cellcolor{teal!41.2}0.412 & \cellcolor{teal!18.3}0.183 & \cellcolor{teal!78.0}0.780 & \cellcolor{teal!69.5}0.695 & \cellcolor{teal!65.2}0.652 & \cellcolor{teal!56.7}0.567 & \cellcolor{teal!47.5}0.475 & \cellcolor{teal!17.6}0.176 & \cellcolor{teal!18.4}0.184 & \cellcolor{teal!7.6}0.076 \\
Qwen-32B-0 & \cellcolor{teal!56.4}0.564 & \cellcolor{teal!35.1}0.351 & \cellcolor{teal!28.6}0.286 & \cellcolor{teal!6.1}0.061 & \cellcolor{teal!79.5}0.795 & \cellcolor{teal!73.2}0.732 & \cellcolor{teal!61.5}0.615 & \cellcolor{teal!51.6}0.516 & - & - & - & - \\
Qwen-72B-0 & - & - & - & - & - & - & - & - & \cellcolor{teal!53.2}0.532 & \cellcolor{teal!26.9}0.269 & \cellcolor{teal!14.8}0.148 & \cellcolor{teal!1.3}0.013 \\
\midrule
Llama-70B-3 & \cellcolor{teal!90.4}0.904 & \cellcolor{teal!84.4}0.844 & \cellcolor{teal!90.5}0.905 & \cellcolor{teal!83.3}0.833 & \cellcolor{teal!83.6}0.836 & \cellcolor{teal!75.2}0.752 & \cellcolor{teal!75.9}0.759 & \cellcolor{teal!68.9}0.689 & \cellcolor{teal!54.2}0.542 & \cellcolor{teal!25.2}0.252 & \cellcolor{teal!61.2}0.612 & \cellcolor{teal!39.5}0.395 \\
Qwen-32B-3 & \cellcolor{teal!92.0}0.920 & \cellcolor{teal!87.5}0.875 & \cellcolor{teal!88.9}0.889 & \cellcolor{teal!85.2}0.852 & \cellcolor{teal!84.1}0.841 & \cellcolor{teal!76.7}0.767 & \cellcolor{teal!69.5}0.695 & \cellcolor{teal!62.8}0.628 & \cellcolor{teal!59.8}0.598 & \cellcolor{teal!30.5}0.305 & \cellcolor{teal!55.0}0.550 & \cellcolor{teal!33.0}0.330 \\
\midrule
Llama-70B-5 & \cellcolor{teal!85.6}0.856 & \cellcolor{teal!77.5}0.775 & \cellcolor{teal!89.2}0.892 & \cellcolor{teal!81.6}0.816 & \cellcolor{teal!84.0}0.840 & \cellcolor{teal!76.3}0.763 & \cellcolor{teal!76.0}0.760 & \cellcolor{teal!68.7}0.687 & \cellcolor{teal!53.9}0.539 & \cellcolor{teal!24.7}0.247 & \cellcolor{teal!56.1}0.561 & \cellcolor{teal!44.5}0.445 \\
Qwen-32B-5 & \cellcolor{teal!92.1}0.921 & \cellcolor{teal!87.9}0.879 & \cellcolor{teal!89.4}0.894 & \cellcolor{teal!85.2}0.852 & \cellcolor{teal!85.0}0.850 & \cellcolor{teal!78.6}0.786 & \cellcolor{teal!69.8}0.698 & \cellcolor{teal!63.5}0.635 & \cellcolor{teal!59.5}0.595 & \cellcolor{teal!30.8}0.308 & \cellcolor{teal!59.3}0.593 & \cellcolor{teal!37.6}0.376 \\
\midrule
Llama-70B-IT & \cellcolor{teal!73.7}0.737 & \cellcolor{teal!60.6}0.606 & \cellcolor{teal!70.2}0.702 & \cellcolor{teal!55.0}0.550 & \cellcolor{teal!82.4}0.824 & \cellcolor{teal!76.1}0.761 & \cellcolor{teal!76.2}0.762 & \cellcolor{teal!70.6}0.706 & \cellcolor{teal!28.4}0.284 & \cellcolor{teal!10.8}0.108 & \cellcolor{teal!77.7}0.777 & \cellcolor{teal!63.0}0.630 \\
Qwen-72B-IT & \cellcolor{teal!71.0}0.710 & \cellcolor{teal!57.0}0.570 & \cellcolor{teal!70.1}0.701 & \cellcolor{teal!53.7}0.537 & \cellcolor{teal!77.1}0.771 & \cellcolor{teal!69.0}0.690 & \cellcolor{teal!73.9}0.739 & \cellcolor{teal!67.5}0.675 & \cellcolor{teal!63.9}0.639 & \cellcolor{teal!36.2}0.362 & \cellcolor{teal!77.5}0.775 & \cellcolor{teal!63.8}0.638 \\

    \bottomrule
    \end{tabular}
    }
    \caption{Combined F1 scores across TBO, ABSA, CSI, and TSD datasets. TF1 = Token-level F1, SF1 = Span-level F1. Here we report the best performing Llama and Qwen model for each approach. Dark Teal cell indicates a higher performance while a lighter tone indicates a lower performance }
    \label{tab:combined-results}
\end{table*}

\paragraph{Evaluation Metrics} We evaluate the performance of the models, using the following two metrics: {\bf Token F1 (TF1)} calculates the F1 score, considering the individual tokens. The final F1-score is the average across all the instances. {\bf Span F1 (SF1)} considers the exact match with the gold standard annotation. The F1 score is calculated considering the total correct predictions across all instances.

\section{Results}
\label{sec:results}

Table \ref{tab:combined-results} show the token-level and span-level F1 scores of the best-performing Llama and Qwen models for all the tasks (the performance of all other models is shown in Table~\ref{tab:all-results}). For TBO and ABSA tasks, few-shot learning achieves the best performance, followed by instruction-tuned models. CoT is the least performing on both these tasks, however, the performance is comparable to zero-shot, in identifying the target and argument spans. However, for ABSA, identifying the aspect spans is more efficient in zero-shot setting compared to CoT.

On simple texts, the instruction-tuned models outperform other approaches for the TSD task. However, instruction-tuned Llama models underperform most other models and approaches for CSI. The 5-shot performance of LLMs for this text type is slightly better than the 3-shot evaluation. In the zero-shot setting, the models achieve a comparable performance to the few-shot evaluation for the CSI, whereas there is a significant performance difference for the TSD. CoT achieves a better performance than the zero-shot for the TSD task, while it underperforms for the CSI task.


\section{Discussion}

In this section, we revisit the four research questions mentioned in Section~\ref{sec:introduction}.

\paragraph{RQ1:} {\bf How does the complexity of the text affect the LLM's ability to identify the different types of spans? Do the models identify specific span types more efficiently than others?} \\

\noindent As seen in Section~\ref{sec:results}, LLMs generally have a better performance on complex spans compared to simple spans. These models are more efficient at identifying the span types that are explicitly mentioned in the text, like targets in TBO or aspects in ABSA. However, LLMs may struggle to identify subjective spans like offensive arguments or opinion terms, that are context-dependent or indirect expressions. For example, the sentence "You are dead to me" may be perceived as offensive, although it does not contain any profane words. Several factors like ambiguity, interpretability, implicit nature, etc, of the spans can influence the model performance, for example in CSI task. While these factors influence the LLM performance in identifying the token spans, several other factors, like identifying irrelevant tokens, splitting the token spans into multiple distinct spans, etc, can also contribute to a lower performance for the LLMs. For example, in TSD task, LLMs tend to identify not only the toxic spans but also the context words, as seen in Figure~\ref{fig:out-tsd}. Such factors especially contribute to the lower Span F1 scores for certain tasks.

\begin{figure*}[!ht]
    \centering
    \begin{subfigure}{0.24\linewidth}
        \includegraphics[width=\linewidth]{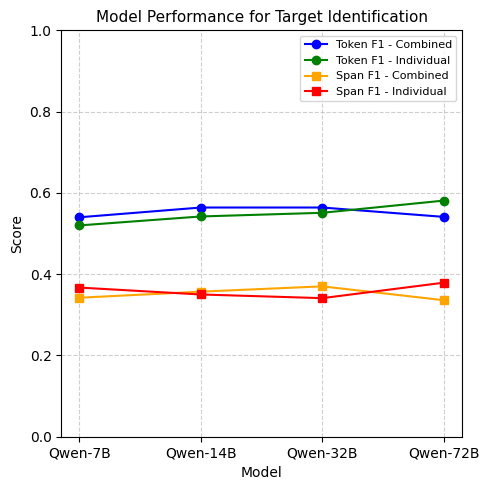}
    \end{subfigure}
    \begin{subfigure}{0.24\linewidth}
        \includegraphics[width=\linewidth]{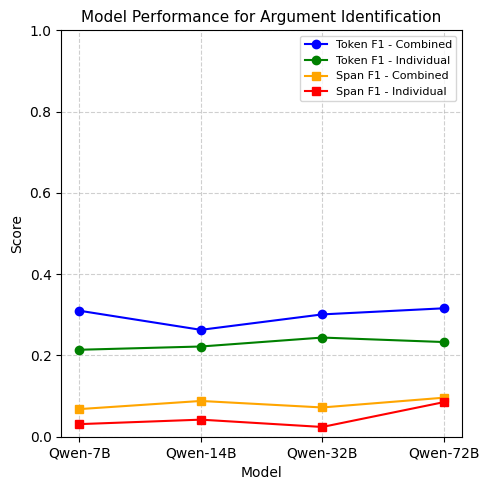}
    \end{subfigure}
    \begin{subfigure}{0.24\linewidth}
        \includegraphics[width=\linewidth]{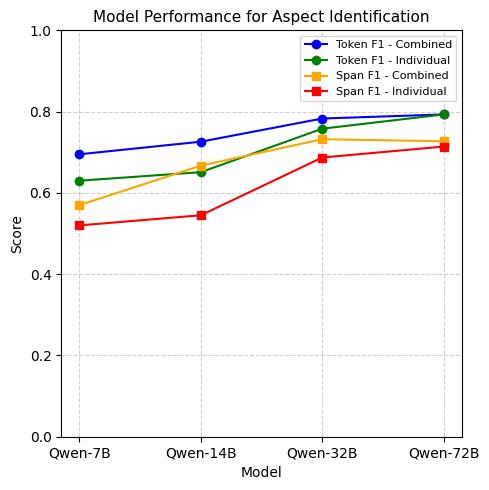}
    \end{subfigure}
    \begin{subfigure}{0.24\linewidth}
        \includegraphics[width=\linewidth]{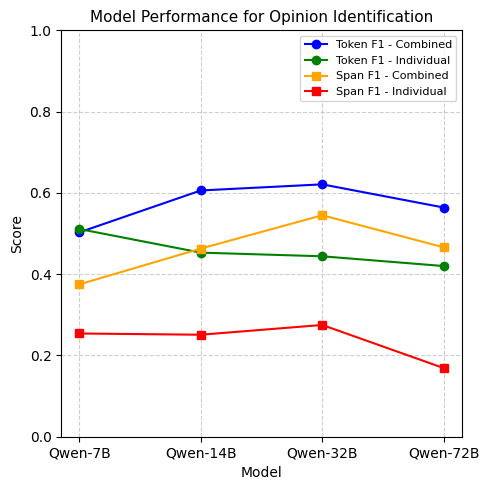}
    \end{subfigure}

    \caption{Token F1 and Span F1 scores for the complex text span identification. The plots show the scores for each type of span when extracted individually and combined.}
    \label{fig:complex-simple}
\end{figure*}

We also aim to identify how text complexity affects the LLMs performance in identifying different types of spans. For the two complex tasks, TBO and ABSA, we assess how well the models identify different span types - individually and combined, using a sample of one hundred instances in a zero-shot setting. As seen in Figure~\ref{fig:complex-simple}, the models, when prompted to identify the two span types together, outperform the models when they are prompted to identify them individually. This indicates that the complexity in the text and the underlying relationships between different span types help the LLMs accurately identify different span types.

\paragraph{RQ2:} {\bf How do the model size and modeling strategies influence the span identification capabilities of LLMs?} \\

\noindent The results indicate that for complex text, LLMs in a few-shot setting outperform all other approaches. In-context examples in the prompts aid the models in identifying the different types of spans. CoT underperforms both zero-shot learning, while IT improves the model performance, especially on subjective spans. Instruction-tuning outperforms all other approaches for both CSI and TSD. The models struggle in zero-shot and CoT for TSD. An analysis of the outputs indicates that some approaches produce irrelevant or extraneous text spans. Furthermore, few-shot learning outperform fine-tuned BERT models on TBO and CSI, while IT has comparable performance to BERT on TSD.

\citet{kaplan2020scaling} show that with increasing model size, the performance improves. To test this hypothesis, we experiment with different model sizes ranging from 7B to 72B parameters. Our experimental results suggest that while the model size increases, there is only a marginal performance improvement. Overall results show that among the Qwen models, the 7B parameter model is the least performing model, whereas among 14B, 32B, and 72B, either model outperforms the others. While for Llama models, the 70B model consistently outperforms the 8B model. However, the performance difference across models with varying parameter sizes is only marginal. With extensive computational resources required for larger models and only marginal performance improvement, smaller models may represent a more efficient choice for these tasks.

\paragraph{RQ3:} {\bf Are LLMs efficient in a low-resource setting?} \\

\noindent Training language models requires extensive training data. However, some of the token classification tasks may have data scarcity. Hence, to assess how the data scarcity affects the performance of language models, fine-tune both small language models (SLM) and large language models (LLM) with varying training data sizes, ranging from 200 to 1000 samples. For this specific experiment, we compare the performance of BERT-large and Qwen-7B models.

Our experiments indicate that, for the TBO, ABSA, and TSD tasks, BERT outperforms Qwen-7B model for all data sizes (See Figure~\ref{fig:progress-test}). The performance varies depending on the span type and number of training examples used, where it is comparable for some span types while substantial for the others. However, the SLM outperforms the LLM on CSI, especially on the span F1 score. This indicates that LLM identifies the exact claim spans more precisely than the smaller models. The findings suggest that SLMs generally outperform LLMs when fine-tuned with limited labeled training data. However, few-shot learning with LLMs can be leveraged in such scenarios due to its higher performance, as indicated in Tables~\ref{tab:combined-results}.


\section{Conclusion}

In this work, we evaluate several LLMs with different approaches on subjective span identification. We answer important research questions pertaining to text complexity and model size, and further explore the capabilities of LLMs in a low-resource setting. Our findings suggest that the complexity and underlying relationships within text aid LLMs in identifying precise text spans. Furthermore, for the specific task of span identification, the model size does not have a significant impact on the performance. Although SLMs like BERT still outperform LLMs, approaches like few-shot learning can be leveraged in a low-resource setting. While LLMs have shown exceptional ability in explicit and context-independent span identification, they still underperform smaller models in identifying subjective spans. 

In future work, we would like to explore approaches to improve the LLMs understanding of the input for accurate span identification, especially considering the context from both left and right. We further plan to explore other challenging datasets on related subjective tasks including multimodal data where text and images are paired \cite{farabi2024survey}. Finally, we plan to expand this work to non-English datasets with the goal of evaluating the multilingual capabilities of the current generation of LLMs. 

\section*{Acknowledgments}

The authors would like to thank the anonymous reviewers for their insightful feedback. 

Marcos Zampieri is partially supported by the Virginia
Commonwealth Cyber Initiative (CCI) award number N-4Q24-009.

\section*{Limitations}

The main limitation of this paper is that we only evaluate two open-source model families. However, other open-source or proprietary models may achieve comparable performance. Additionally, the prompts used in the experiments follow a specific template style. Experimenting with different prompt templates may generate different results. The task-specific instructions within the prompts can be adjusted to generate more efficient outputs. In the few-shot experiments, we use three and five examples in the prompts. However, including additional examples can enhance the model's performance. Moreover, our evaluations are focused on English datasets. Expanding this work to encompass additional languages and task datasets may offer further insights into the token span identification capabilities of LLMs.

\bibliography{custom}

\newpage

\appendix

\section{Appendix}
\label{sec:appendix}

\subsection{LLM Prompt}\label{sec:inst-prompts}
Figure~\ref{fig:prompt} shows the prompt template used in our experiments. Table~\ref{tab:inst-tune-prompts} shows the task-specific prompts. Each prompt consists of instructions describing the tasks and each type of span to extract. Additionally, the prompts include the input instances.

\begin{figure}[!ht]
    \centering
    \noindent\fbox{%
        \parbox{0.95\linewidth}{%
            \textit{<Task-specific instruction. This includes the definitions of the type of spans to be identified>\\ \\
    {\bf Output Format:} \\
    <The format of the generated output. This will be 'tag: span'. For example, Target: target-span> \\ \\
    {\bf Examples:} \\
    <n examples> \\ \\
    {\bf Input:} \\
    <Input Text> \\ \\
    {\bf Response}:
        }
    }
    }
    \caption{The prompt template used in our experiments.}
    \label{fig:prompt}
\end{figure}

\begin{table*}[!ht]
    \centering
    \resizebox{0.92\textwidth}{!}{
    \begin{tabular}{cp{15.5cm}}
    \toprule
         {\bf Dataset} & {\bf Prompt} \\
         \midrule
         \multirow{10}{*}{TBO} & {For the given text, identify the (target, argument) pairs.

• Target: The individual, group, or organization towards whom the argument is directed.

• Argument: A phrase or sentence containing offensive, profane, or unacceptable language.

• An argument may or may not have a target.

• Target can appear more than once if referenced by multiple arguments.

• The argument and target may be the same.

\bigbreak
Output Format:

Target n: <nth target>

Argument n: <nth argument>

\bigbreak
Examples:\{n examples\}

\bigbreak
Input:

\{input text\}

\bigbreak
Response:}\\

         \midrule
         \multirow{8}{*}{ABSA} & {For the given text, identify the aspects and opinions.

• Aspect: The entities to which sentiments are tied to.

• Opinion: The sentiment words or phrases.

\bigbreak
Output Format:

Aspect n: <nth aspect>

Opinion n: <nth opinion>

\bigbreak
Examples:\{n examples\}

\bigbreak
Input:

\{input text\}

\bigbreak
Response:}\\
         \midrule
         \multirow{8}{*}{CSI} & {For the given text, identify the claim spans.

• Claim Span: A phrase or sentence that explicitly mentions a claim, assertion, or argument.

\bigbreak
Output Format:

Span n: <nth claim span>

\bigbreak
Examples:\{n examples\}

\bigbreak
Input:

\{input text\}

\bigbreak
Response:}\\
\midrule
         \multirow{8}{*}{TSD} & {For the given text, identify the arguments.

• Argument: A phrase or sentence containing offensive, profane, or unacceptable language.

\bigbreak
Output Format:

Argument n: <nth argument>

\bigbreak
Examples:\{n examples\}

\bigbreak
Input:

\{input text\}

\bigbreak
Response:} \\

         \bottomrule
    \end{tabular}}
    \caption{Prompts used for each task. The prompt contains a task-specific instruction along with the respective input instance.}
    \label{tab:inst-tune-prompts}
\end{table*}

\subsection{Hyper-parameters}
To fine-tune the models, we experiment with several learning rate values, with 1e-4 giving an optimal performance and minimum average loss. A per-device batch size of 2 with a gradient accumulation size of 8 was used to instruction-tune the LLMs. We chose a lower batch size to accommodate the limited computational resources available. We further leveraged the Adam optimizer and fine-tuned the models for ten epochs. For LoRA, we use the alpha value of 16 and the r value of 64, as these values provided the best performance. A dropout of 0.1 was used. To evaluate the LLMs with CoT and ICL, we use a temperature value of 0.0001 (as models like Llama do not allow a temperature value of 0) to allow deterministic outputs.

\subsection{Additional Results}
In this section, we present additional results for our experiments. While Table~\ref{tab:combined-results} shows the F1-scores of the best performing Qwen and Llama models, we present the results for all other models in Table~\ref{tab:all-results}.

\begin{table*}[!ht]
    \centering
    \resizebox{\textwidth}{!}{
    \begin{tabular}{lcccccccc|cccc}
    \toprule
        \multirow{3}{*}{\bf Model} 
        & \multicolumn{4}{c}{\bf TBO} 
        & \multicolumn{4}{c|}{\bf ABSA}  
        & \multicolumn{2}{c}{\multirow{2}{*}{\bf CSI}} 
        & \multicolumn{2}{c}{\multirow{2}{*}{\bf TSD}} \\
        \cmidrule{2-9}
        & \multicolumn{2}{c}{\bf Target} & \multicolumn{2}{c}{\bf Argument} 
        & \multicolumn{2}{c}{\bf Aspect} & \multicolumn{2}{c|}{\bf Opinion}  
        &  &  &  & \\
        \cmidrule{2-13}
        & TF1 & SF1 & TF1 & SF1 & TF1 & SF1 & TF1 & SF1 
        & TF1 & SF1 & TF1 & SF1 \\
        \midrule
        Llama-8B-CoT & 0.533 & 0.171 & 0.362 & 0.052 & 0.569 & 0.333 & 0.558 & 0.236 & 0.203 & 0.043 & 0.515 & 0.012 \\
         Qwen-7B-CoT & 0.518 & 0.282 & 0.296 & 0.060 & 0.649 & 0.509 & 0.507 & 0.340 & 0.429 & 0.210 & 0.262 & 0.029 \\
         Qwen-14B-CoT & 0.541 & 0.151 & 0.329 & 0.017 & 0.607 & 0.457 & 0.582 & 0.410 & 0.443 & 0.188 & 0.323 & 0.019 \\
         Qwen-32B-CoT & 0.578 & 0.329 & 0.316 & 0.056 & 0.576 & 0.360 & 0.613 & 0.302 & 0.444 & 0.169 & 0.425 & 0.025 \\
         \midrule
         Llama-8B-0 & 0.516 & 0.310 & 0.225 & 0.044 & 0.707 & 0.574 & 0.524 & 0.401 & 0.469 & 0.198 & 0.135 & 0.036 \\
         Qwen-7B-0 & 0.531 & 0.314 & 0.302 & 0.078 & 0.721 & 0.606 & 0.526 & 0.404 & 0.517 & 0.275 & 0.321 & 0.024 \\
         Qwen-14B-0 & 0.563 & 0.352 & 0.255 & 0.048 & 0.726 & 0.648 & 0.587 & 0.436 & 0.516 & 0.227 & 0.181 & 0.010 \\
         Qwen-32B-0 & - & - & - & - & - & - & - & - & 0.528 & 0.223 & 0.253 & 0.008 \\
         Qwen-72B-0 & 0.544 & 0.332 & 0.313 & 0.090 & 0.776 & 0.705 & 0.564 & 0.462 & - & - & - & - \\
         \midrule
         Llama-8B-3 & 0.880 & 0.792 & 0.881 & 0.786 & 0.803 & 0.681 & 0.700 & 0.577 & 0.503 & 0.248 & 0.559 & 0.299 \\
         Qwen-7B-3 & 0.898 & 0.834 & 0.881 & 0.812 & 0.791 & 0.709 & 0.628 & 0.563 & 0.539 & 0.275 & 0.444 & 0.244 \\
         Qwen-14B-3 & 0.913 & 0.864 & 0.897 & 0.848 & 0.816 & 0.746 & 0.662 & 0.586 & 0.582 & 0.295 & 0.542 & 0.302 \\
         Qwen-72B-3 & 0.852 & 0.791 & 0.873 & 0.812 & 0.835 & 0.779 & 0.686 & 0.627 & 0.562 & 0.281 & 0.550 & 0.343 \\
         \midrule
         Llama-8B-5 & 0.836 & 0.735 & 0.874 & 0.759 & 0.801 & 0.680 & 0.716 & 0.592 & 0.495 & 0.228 & 0.597 & 0.342 \\
         Qwen-7B-5 & 0.899 & 0.840 & 0.889 & 0.824 & 0.804 & 0.728 & 0.645 & 0.586 & 0.530 & 0.266 & 0.502 & 0.290 \\
         Qwen-14B-5 & 0.880 & 0.816 & 0.887 & 0.824 & 0.830 & 0.761 & 0.671 & 0.603 & 0.587 & 0.292 & 0.582 & 0.349 \\
         Qwen-72B-5 & 0.838 & 0.776 & 0.868 & 0.801 & 0.845 & 0.789 & 0.708 & 0.645 & 0.573 & 0.277 & 0.586 & 0.388 \\
         \midrule
         Llama-8B-IT & 0.707 & 0.539 & 0.692 & 0.537 & 0.790 & 0.717 & 0.755 & 0.670 & 0.178 & 0.000 & 0.768 & 0.629 \\
         Qwen-7B-IT & 0.645 & 0.392 & 0.644 & 0.441 & 0.730 & 0.619 & 0.571 & 0.471 & 0.592 & 0.316 & 0.747 & 0.590 \\
         Qwen-14B-IT & 0.678 & 0.482 & 0.673 & 0.514 & 0.776 & 0.689 & 0.715 & 0.643 & 0.612 & 0.334 & 0.753 & 0.589 \\
         Qwen-32B-IT & 0.669 & 0.475 & 0.654 & 0.485 & 0.783 & 0.712 & 0.700 & 0.646 & 0.590 & 0.319 & 0.756 & 0.600 \\
        \bottomrule
    \end{tabular}
    }
    \caption{F1 scores of 0-, 3-, 5-shot, CoT, and instruction tuned (IT) models. The TF1 and SF1 scores for each span type are reported.}
    \label{tab:all-results}
\end{table*}

\subsection{Task-Specific Outputs}
We show outputs for all the tasks in Figures~\ref{fig:out-tbo}-~\ref{fig:out-tsd}. For each task, the outputs generated by the Llama model with different approaches are shown. For the few-shot setting, we show the outputs of 5-shot experiments.

\begin{figure*}[!ht]
    \centering
    \noindent\fbox{%
        \parbox{0.95\linewidth}{%
            {\bf Input:} I'll be tweeting utter bullshit and sheer nonsense at this time of the night \\
    \rule{\linewidth}{0.4pt} 
    {\bf 0-Shot:} \\
    Target 1: I\\Argument 1: utter bullshit and sheer nonsense \\
    \rule{\linewidth}{0.4pt} 
    {\bf 5-Shot:} \\
    Target 1: nan\\Argument 1: utter bullshit\\Target 2: nan\\Argument 2: sheer nonsense \\
    \rule{\linewidth}{0.4pt} 
    {\bf Instruction Tuned:} \\
    Target 1: None\\Argument 1: utter bullshit\\Target 2: None\\Argument 2: sheer nonsense \\
    \rule{\linewidth}{0.4pt} 
    {\bf Chain of Thought:} \\
    Let's think step by step. The text contains one argument: "utter bullshit and sheer nonsense". The target of this argument is not explicitly mentioned, but it can be inferred that the target is the person or people who are being addressed by the speaker. Therefore, the output would be:\\Target 1: The person or people being addressed by the speaker\\Argument 1: utter bullshit and sheer nonsense
        }
    }
    \caption{Outputs for the TBO task. The output for each approach is shown. For few-shot, only the 5-shot output is shown.}
    \label{fig:out-tbo}
\end{figure*}

\begin{figure*}[!ht]
    \centering
    \noindent\fbox{%
        \parbox{0.95\linewidth}{%
            {\bf Input:} after dinner i heard music playing and discovered that there is a lounge downstairs \\
    \rule{\linewidth}{0.4pt} 
    {\bf 0-Shot:} \\
    Aspect 1: music\\Opinion 1: playing\\Aspect 2: lounge\\Opinion 2: downstairs \\
    \rule{\linewidth}{0.4pt} 
    {\bf 5-Shot:} \\
    Aspect 1: music\\Aspect 2: lounge\\Opinion 1: playing\\Aspect 3: downstairs \\
    \rule{\linewidth}{0.4pt} 
    {\bf Instruction Tuned:} \\
    Aspect 1: music\\Opinion 1: playing\\Aspect 2: lounge\\Opinion 2: downstairs \\
    \rule{\linewidth}{0.4pt} 
    {\bf Chain of Thought:} \\
    Let's think step by step. First, we need to identify the aspects in the given text. In this case, the aspect is "music playing" and "lounge downstairs". \\Next, we need to identify the opinions in the text. The opinion is "I heard music playing" and "I discovered that there is a lounge downstairs". \\So, the output format would be:\\Aspect 1: music playing\\Opinion 1: I heard music playing\\Aspect 2: lounge downstairs\\Opinion 2: I discovered that there is a lounge downstairs
        }
    }
    \caption{Outputs for the ABSA task. The output for each approach is shown. For few-shot, only the 5-shot output is shown.}
    \label{fig:out-absa}
\end{figure*}

\begin{figure*}[!ht]
    \centering
    \noindent\fbox{%
        \parbox{0.95\linewidth}{%
            {\bf Input:} Getting the jab protects you from hurricanes? uhh.. let’s go Brandon! \\
    \rule{\linewidth}{0.4pt} 
    {\bf 0-Shot:} \\
    Span 1: Getting the jab protects you from hurricanes?\\Span 2: uhh.. let's go Brandon! \\
    \rule{\linewidth}{0.4pt} 
    {\bf 5-Shot:} \\
    Span 1: Getting the jab protects you from hurricanes?\\Span 2: uhh.. let's go Brandon! \\
    \rule{\linewidth}{0.4pt} 
    {\bf Instruction Tuned:} \\
    Span 1: Getting the jab protects you from hurricanes?\\Span 2: uhh.. let's go Brandon! \\
    \rule{\linewidth}{0.4pt} 
    {\bf Chain of Thought:} \\
    Let's think step by step. The text contains one claim span. It is the phrase "Getting the jab protects you from hurricanes?" which is an assertion or argument made by the speaker. Therefore, the output would be:\\Span 1: Getting the jab protects you from hurricanes?
        }
    }
    \caption{Outputs for the CSI task. The output for each approach is shown. For few-shot, only the 5-shot output is shown.}
    \label{fig:out-csi}
\end{figure*}

\begin{figure*}[!ht]
    \centering
    \noindent\fbox{%
        \parbox{0.95\linewidth}{%
            {\bf Input:} This proposal is so idiotic that its failure is imminent. \\
    \rule{\linewidth}{0.4pt} 
    {\bf 0-Shot:} \\
    Argument 1: This proposal is so idiotic that its failure is imminent. \\
    \rule{\linewidth}{0.4pt} 
    {\bf 5-Shot:} \\
    Argument 1: idiotic \\
    \rule{\linewidth}{0.4pt} 
    {\bf Instruction Tuned:} \\
    Argument 1: This proposal is so idiotic that its failure is imminent. \\
    \rule{\linewidth}{0.4pt} 
    {\bf Chain of Thought:} \\
    Let's think step by step. The given text is: "This proposal is so idiotic that its failure is imminent."\\Step 1: Identify the phrases or sentences that contain offensive, profane, or unacceptable language.\\- In this case, the phrase "idiotic" can be considered offensive or unacceptable language.\\Step 2: Determine the arguments based on the identified phrases or sentences.\\- Argument 1: "idiotic"\\Output Format:\\Argument 1: idiotic
        }
    }
    \caption{Outputs for the TSD task. The output for each approach is shown. For few-shot, only the 5-shot output is shown.}
    \label{fig:out-tsd}
\end{figure*}



\subsection{Progress Test}
\begin{figure*}[!ht]
    \centering
    \begin{subfigure}{0.47\linewidth}
        \includegraphics[width=\linewidth]{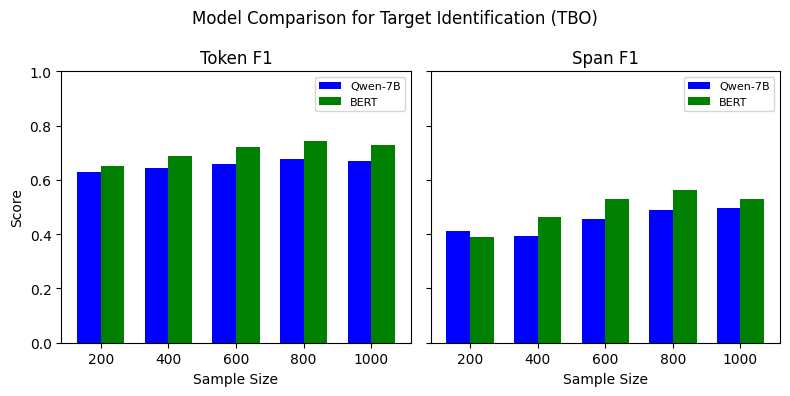}
    \end{subfigure}
    \begin{subfigure}{0.47\linewidth}
        \includegraphics[width=\linewidth]{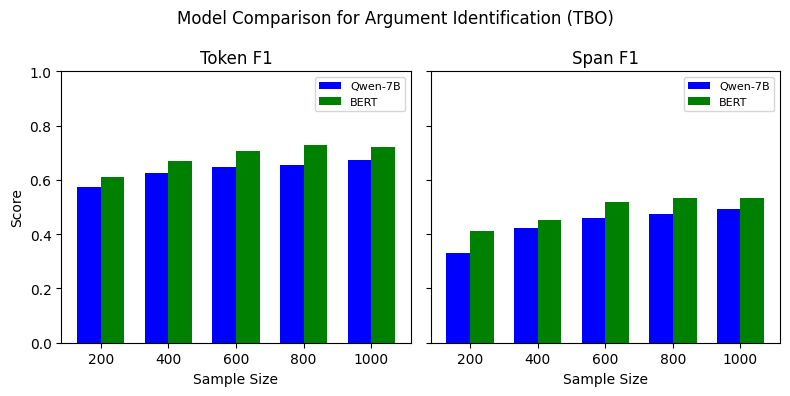}
    \end{subfigure}
    \begin{subfigure}{0.47\linewidth}
        \includegraphics[width=\linewidth]{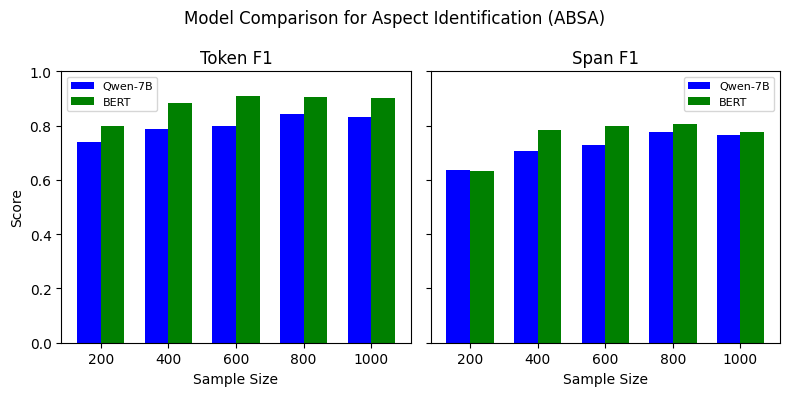}
    \end{subfigure}
    \begin{subfigure}{0.47\linewidth}
        \includegraphics[width=\linewidth]{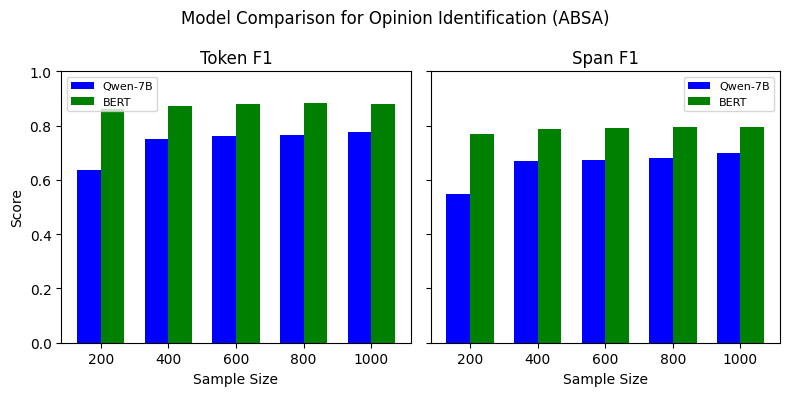}
    \end{subfigure}
    \begin{subfigure}{0.47\linewidth}
        \includegraphics[width=\linewidth]{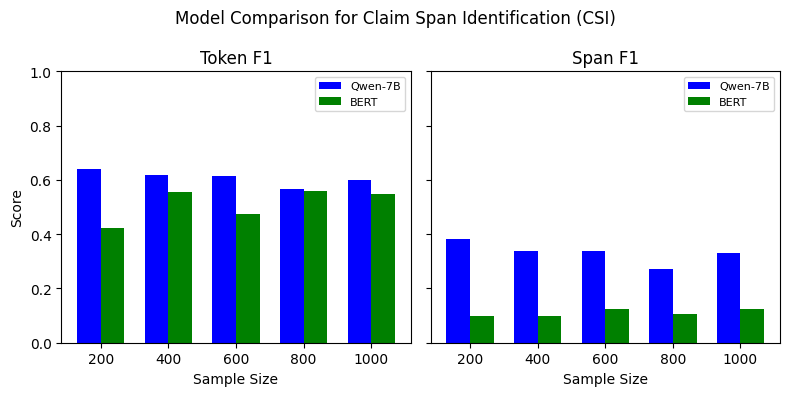}
    \end{subfigure}
    \begin{subfigure}{0.47\linewidth}
        \includegraphics[width=\linewidth]{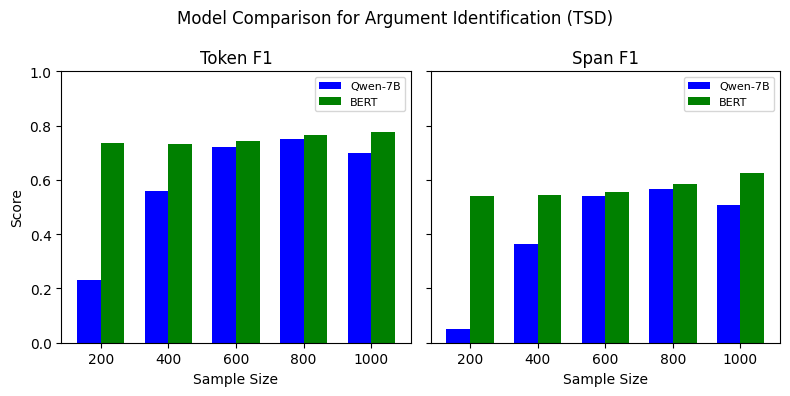}
    \end{subfigure}

    \caption{Progress test results. For each span type, the TF1 and SF1 scores are reported}
    \label{fig:progress-test}
\end{figure*}

Figure~\ref{fig:progress-test} shows the performance of Qwen-7B compared to BERT when trained with varying training data sizes. We perform this experiment to understand how LLMs perform in a low-resource setting. For the TBO and ABSA tasks, BERT outperform the Qwen-7B model. Similarly, for TSD, BERT outperforms Qwen-7B when fine-tuned with 200 instances, but as we increase the number of training instances, the F1 score difference decreases. Unlike the other three tasks, Qwen-7B outperforms BERT, with a substantial Span F1 difference between the two models. However, as we increase the training dataset, the difference in Token F1 for this task gradually decreases. The experiment indicates that the performance of SLMs and LLMs differs, considering the type of task.

\end{document}